\documentclass{article}

\usepackage[nonatbib, final]{neurips_2023_ml4ps}

\usepackage[utf8]{inputenc} % allow utf-8 input
\usepackage[T1]{fontenc}    % use 8-bit T1 fonts
\usepackage{hyperref}       % hyperlinks
\usepackage{url}            % simple URL typesetting
\usepackage{booktabs}       % professional-quality tables
\usepackage{amsfonts}       % blackboard math symbols
\usepackage{nicefrac}       % compact symbols for 1/2, etc.
\usepackage{microtype}      % microtypography
\usepackage{xcolor}         % colors

\usepackage{graphicx}
\usepackage{amssymb}
\usepackage{amsmath}
\usepackage{caption} % subfigures
\usepackage{subcaption} % subfigures
\usepackage{fontawesome5} % for github logo

\newcommand{\github}[1]{%
   \href{#1}{\textcolor{black}\faGithubSquare}%
}
\def\ltsima{$\; \buildrel < \over \sim \;$}
\def\simlt{\lower.5ex\hbox{\ltsima}}
\def\gtsima{$\; \buildrel > \over \sim \;$}
\def\simgt{\lower.5ex\hbox{\gtsima}}
\def\s{\ifmmode \widetilde \else \~\fi}
\def\={\overline}

\def\spose#1{\hbox to 0pt{#1\hss}}

\def\eg{{e.g.,\ }}
\def\ie{{i.e.\ }}
\def\lta{\mathrel{\spose{\lower 3pt\hbox{$\mathchar"218$}}
     \raise 2.0pt\hbox{$\mathchar"13C$}}}
\def\gta{\mathrel{\spose{\lower 3pt\hbox{$\mathchar"218$}}
     \raise 2.0pt\hbox{$\mathchar"13E$}}}
\def\Dt{\spose{\raise 1.5ex\hbox{\hskip3pt$\mathchar"201$}}}    % upper case
\def\dt{\spose{\raise 1.0ex\hbox{\hskip2pt$\mathchar"201$}}}    % lower case

\def\dotsfill{\leaders\hbox to 1em{\hss.\hss}\hfill}

\def\ltsima{$\; \buildrel < \over \sim \;$}
\def\gtsima{$\; \buildrel > \over \sim \;$}
\def\lsim{\lower.5ex\hbox{\ltsima}}
\def\gsim{\lower.5ex\hbox{\gtsima}}
\def\lapp{\ifmmode\stackrel{<}{_{\sim}}\else$\stackrel{<}{_{\sim}}$\fi}
\def\gapp{\ifmmode\stackrel{>}{_{\sim}}\else$\stackrel{<}{_{\sim}}$\fi}

\graphicspath{{./}{figures/}}

\def\PhySO{{$\Phi$-SO}}
\def\insitu{{\textit{in situ}\ }}
\def\Insitu{{\textit{In situ}\ }}

\def\placeholder{{\square}}

\title{Physical Symbolic Optimization}

\author{%
  Wassim Tenachi\\
  Universit\'e de Strasbourg, CNRS\\
  Observatoire astronomique de Strasbourg\\
  UMR 7550, F-67000 Strasbourg, France \\
  \texttt{wassim.tenachi@astro.unistra.fr} \\
  \And
  Rodrigo Ibata\\
  Universit\'e de Strasbourg, CNRS\\
  Observatoire astronomique de Strasbourg\\
  UMR 7550, F-67000 Strasbourg, France \\
  \texttt{rodrigo.ibata@astro.unistra.fr} \\
  \And
  Foivos I. Diakogiannis\\
  Data61, CSIRO\\
  Kensington, WA 6155, Australia\\
  \texttt{foivos.diakogiannis@data61.csiro.au} \\
}

\begin{document}

\maketitle

\begin{abstract}
We present a framework for constraining the automatic sequential generation of equations to obey the rules of dimensional analysis by construction. Combining this approach with reinforcement learning, we built $\Phi$-SO, a Physical Symbolic Optimization method for recovering analytical functions from physical data leveraging units constraints. Our symbolic regression algorithm achieves state-of-the-art results in contexts in which variables and constants have known physical units, outperforming all other methods on SRBench's Feynman benchmark in the presence of noise (exceeding 0.1\%) and showing resilience even in the presence of significant (10\%) levels of noise.
\end{abstract}

\section{Introduction}
\label{sec:intro}

Physical theories traditionally stem from empirical laws. Physicists typically observe natural phenomena, formulate empirical laws to describe them, and subsequently construct overarching theories that encompass these laws. For example, Newton's law of universal gravitation was designed to explain both the motion of terrestrial objects as well as Kepler's law of planetary motion \cite{NewtonPrincipia}. However, with the rise of deep learning, many empirical laws have transitioned into complex neural network representations, rendering their integration into overarching physical theories expressed as analytical equations much more complex. Symbolic regression (SR) emerges as a pivotal approach to bridge this gap in interpretability.

\paragraph{Symbolic regression}
SR consists in the inference of a symbolic analytical function $f: \mathbb{R}^n \longrightarrow \mathbb{R}$ that accurately represents the relationship $y = f(\mathbf{x})$ provided a data pair $(\mathbf{x}, y)$. Unlike numerical parameter optimization methods, SR involves the exploration of the space of functional forms themselves by optimizing the arrangement of mathematical symbols (such as $x$, $+$, $-$, $\times$, $/$, $\sin$, $\exp$, $\log$, and more). The discrete combinatorial challenge presented by SR makes it an ``NP hard'' (nondeterministic polynomial time) problem \cite{SRisNPhard} without even factoring the additional challenge of optimizing free constants in a continuous space.
SR therefore requires the development of efficient strategies for avoiding sub-optimal guesses.

\paragraph{Related works}
SR has traditionally been tackled using genetic programming, and is notably at the heart of the \texttt{Eureqa} software \cite{EureqaPaper2009} and its successor algorithm \texttt{AFP\_FE} \cite{EureqaPaper2011_AFP}. However, with the advance of deep learning techniques, multiple frameworks have been proposed to use neural networks for SR \cite{SRBench}.
% Generating expressions with an RNN
As in previous deep learning SR work \cite{Kamienny_EndToEndSR, NeSymReS_EndToEndSR, PetersenDSR}, here we employ a recurrent neural network (RNN) to repeatedly sample tokens representing mathematical symbols or variables: ultimately generating complete expressions in prefix notation\footnote{There is a one-to-one relation between this representation which is commonly used in symbolic computation and the ``infix'' representation we are more familiar with.} in which operators are written first and their arguments second, which alleviates the need for parentheses.
% Reinforcement learning
This type of approach includes \texttt{DSR} \cite{PetersenDSR, SR_PG_improvements} which is the current state-of-the-art approach for exact symbolic recovery in the presence of noise \cite{SRBench}. This approach relies on a risk-seeking reinforcement learning policy, which we adopt in this study.
% Previous dimensional analysis approaches
We also note that in the context of the physical sciences, previous works (\eg \cite{AIFeynman2,SR_appli_exoplanet_units}) have taken into consideration the units associated with data, proposing to render datasets dimensionless using the Buckingham $\Pi$ theorem \cite{Buckingham_pi_theorem}. This effectively renders variables and constants dimensionless by means of multiplicative operations amongst them as an initial step before SR, thereby relinquishing symbolic arrangement constraints arising from dimensional analysis but ensuring the physical consistency of generated expressions in terms of units.

\paragraph{Contribution}
Here we propose a Physical Symbolic Optimization (\PhySO) framework. This framework operates in contexts in which the units of variables and constants involved are known leveraging dimensional analysis to constrain the symbolic arrangement during the generation of equations. Specifically, \PhySO\ builds upon the reinforcement learning SR framework pioneered in \cite{PetersenDSR}, equipping it with an \insitu supervised learning component designed to teach the RNN dimensional analysis rules, guiding its exploration of the search space towards not only accurate but physically meaningful combinations.

In Section \ref{sec:method}, we detail our methodology. In Section \ref{sec:results}, we show that in physical contexts in which all units involved are known, \PhySO\ outperforms state-of-the-art baselines in exact symbolic recovery and discuss those results. Finally, Section \ref{sec:conclusion} summarizes our conclusions.

\section{Method}
\label{sec:method}

\begin{figure*}[h]
\begin{center}
\includegraphics[angle=0, clip, width=\hsize]{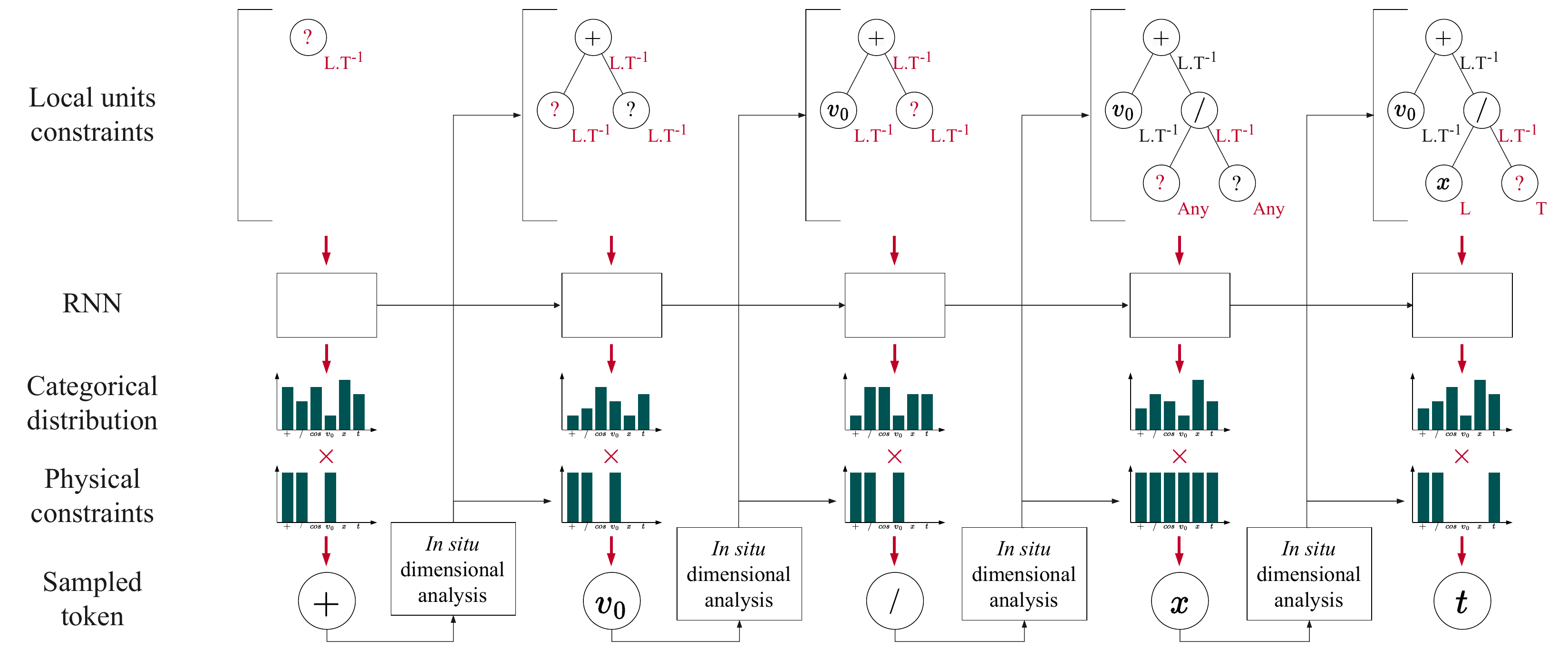}
\end{center}
\caption{Expression generation sketch. For each token, the RNN is given symbolic and units related contextual information regarding the next token to generate. Based on this information, the RNN produces a categorical distribution over the space of available tokens (top histograms) as well as a memory state which is given to the RNN on its next call. \PhySO\ generates \insitu categorical labels encoding which tokens can be chosen based on dimensional analysis rules (bottom histograms). The generated distribution is then masked based on those constraints, forbidding tokens that would lead to nonsensical expressions. The resulting token is sampled from this distribution, \PhySO\ then updates the local unit constraints of the partial tree representing the expression being generated. Repeating this process, from left to right, allows one to generate a complete physical expression, here $[+, v_0, /, x, t]$ which translates into $v_0+x/t$ in the infix notation we are more familiar with.}
\label{fig:embedding}
\end{figure*}

\paragraph{Expression generation}
The expression generation procedure is detailed in Figure \ref{fig:embedding}. At each token generation step, the RNN, here a long-short term memory (LSTM) type RNN \cite{LSTM}, is given as input contextual information regarding the next token to be generated. This includes the nature of its parent token in a tree representation of the expression as well as its units or local units constraints if the token is a mathematical operation. Similar information is passed regarding the sibling and the previously generated token. In addition, the required units of the token to be generated and the number of dangling nodes in a tree representation (\ie number of tokens required to finish the expression) are also passed to the RNN. This information is then leveraged by the RNN to produce a categorical distribution over the space of choosable tokens maximizing the probabilities of tokens obeying dimensional analysis rules and resulting in accurate expressions.

\paragraph{\Insitu dimensional analysis}
As illustrated in Figure \ref{fig:embedding} before each token generation step, we run an algorithm that updates the local units constraints\footnote{Since the algorithm only has access to an incomplete expression containing dangling nodes, it is sometimes impossible to compute the required units of a token, in these cases, the token units are temporarily marked as free \ie able to accommodate any units.} of the expression nodes based on the new token sampled at the previous step following the dimensional analysis prescriptions given in Table \ref{table:dimensional_analysis}. This information is used to produce the units related inputs of the RNN as well as to constrain its output categorical distribution in favor of tokens obeying dimensional analysis. Effectively, we deterministically produce inputs and labels on the fly for the RNN to learn on in a supervised manner.

\newcommand{\ope}{\textcolor{black}{\text{op}_\mathbf{0}}}
\newcommand{\varA}{{\tau_A}}
\newcommand{\varB}{{\tau_B}}
\newcommand{\vary}{{y}}
\newcommand{\uA}{{\Phi_A}}
\newcommand{\uB}{{\Phi_B}}
\newcommand{\uy}{{\Phi_y}}

\begin{table}[h]
    \begin{subtable}[h]{0.45\textwidth}
        \centering
        
       \begin{tabular}{cc}
       Expression           & Units               \\ \hline
       $\varA \pm \varB$    & $\uA$ or  $\uB$     \\
       $-\varA$             & $\uA$               \\
       $\varA \times \varB$ & $\uA + \uB$         \\
       $\varA / \varB$      & $\uA - \uB$         \\
       $\varA^n$            & $n \times \uA$      \\
       $\ope$$(\varA)$      & $\mathbf{0}$        \\ \hline
       \end{tabular}
       
       \caption{Dimensional analysis rules}
       \label{subtable:DA_rules}
    \end{subtable}
    \hfill
    \begin{subtable}[h]{0.45\textwidth}
        \centering
        \begin{tabular}{cc}
        Expression               & Requirement       \\ \hline
        $\varA \pm \varB$        & $\uA = \uB$        \\
        $\vary = \varA$          & $\uy = \uA$ \\
        $\ope$$(\varA)$          & $\uA = \mathbf{0}$ \\ 
        \hline
        \end{tabular}
        \caption{Units requirements rules}
        \label{subtable:units_requirements_rules}
     \end{subtable}
     \caption{Dimensional analysis prescriptions to enforce. With $\varA$, $\varB$, $y$, $\uA$, $\uB$, $\uy$ referring to two nodes, the output variable and the powers of their units vectors, $\ope$ denoting a dimensionless operation (\eg $\{\cos,\sin, \exp, \log \}$) and $\varA^n$ representing any power operation (including \eg $1/\varA = \varA^{-1}$, $\sqrt{\varA} = \varA^{\frac{1}{2}}$)}
     \label{table:dimensional_analysis}
\end{table}

\paragraph{Training}
Expressions are rewarded based on their fit quality on a given target dataset $(\mathbf{x}, y)$. For each expression, free constants appearing in it are first optimized using the LBFGS \cite{LBFGS} algorithm and the reward defined as the squashed normalized root mean squared error ($\text{Reward} = 1/(1+\text{NRMSE})$) is then computed. Subsequently, the RNN is trained based on those rewards following the risk-seeking policy gradients detailed in \cite{PetersenDSR} with improvements from \cite{SR_PG_improvements}. It should be noted that in order to prevent the expression generation phase going on forever, in addition to dimensional analysis considerations, the categorical distribution emitted by the RNN is masked in an expression length dependent manner such as to favor terminal nodes (\ie variables and constants), thereby encouraging its termination, before a maximum expression size limit is reached. This indispensable length prior can conflict with the constraints arising from our dimensional analysis module, resulting in the expression being discarded. This conflict typically results in $\sim 90\%$ of expressions being discarded in the first few training epochs. This makes it essential for the RNN to learn dimensional analysis rules and not solely rely on the deterministic constraints imposed \textit{a posteriori} in order to avoid such situations. This learning process can typically be monitored by keeping track of the rate of discarded expressions as a function of training epochs.

\section{Results \& Discussion}
\label{sec:results}

We evaluated \PhySO\ following the \texttt{SRBench} standard SR benchmarking procedure \cite{SRBench} (\href{https://github.com/cavalab/srbench}{github.com/cavalab/srbench}) against 17 baseline SR methods. Conveniently, the bulk of this benchmark consists of 116 ground-truth challenges from \cite{AIFeynman} with physical units associated with each variable involved making it straightforward to apply our method.
It is worth noting that, despite the availability of units in this benchmark, \PhySO\ and \texttt{AI Feynman} \cite{AIFeynman} are the only methods that have harnessed this dataset to leverage the associated units.

\begin{figure*}
\begin{center}
\includegraphics[angle=0, clip, width=\hsize]{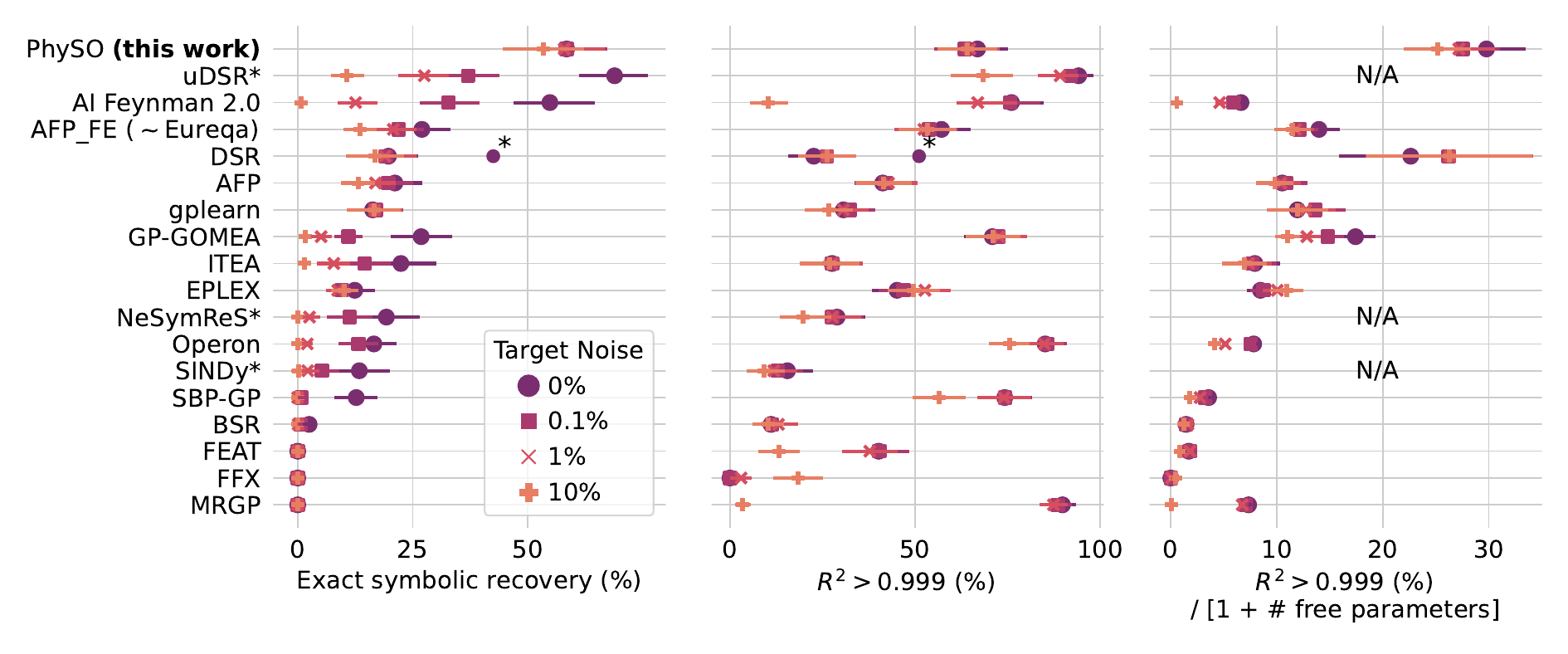}
\caption{Exact symbolic recovery rate, rates of accurate expression (having $R^2>0.999$) and rate of accurate expressions normalized by the number of free parameters appearing in expressions for our \texttt{PhySO} algorithm and 17 baselines, averaged across 116 ground-truth Feynman SR problems following the standardized \texttt{SRBench} \cite{SRBench} benchmarking method.* denotes results taken from \cite{uDSR_DSRbased}.}
\label{fig:feynman_benchmark}
\end{center}
\end{figure*}

We maintained strict consistency by employing the same set of hyper-parameters for all challenges. These parameters included permitting the use of $\{ +, -, \times, /, 1/\placeholder, \sqrt{\placeholder}, \placeholder^2, -{\placeholder}, \exp, \log, \cos, \sin\}$\ as well as two dimensionless adjustable free constants and a constant equal to one $\{\theta_1, \theta_2, 1\}$.
In addition, we strictly adhered to the established benchmarking rules, which included employing a dataset consisting of only 10,000 data points and limiting the maximum number of expression evaluations during the search to 1 million for each challenge. This computational process typically takes about 1 hour, utilizing all cores of an Intel Xeon W-2155 CPU. It should be noted that \PhySO\ typically converges toward the correct expression well before reaching this evaluation limit or not at all.
\PhySO's performances on these challenges against baseline methods are given in Figure \ref{fig:feynman_benchmark}.

\definecolor{NoiselessViolet}{HTML}{7a2e70}
\def\properDSR{{\textcolor{NoiselessViolet}{\blacklozenge}}}

\paragraph{Exact symbolic recovery}
% Improvement over DSR
In contrast to \texttt{DSR}, which relies solely on reinforcement learning, \PhySO\ leverages both reinforcement learning and dimensional analysis. This combined approach yields significantly improved results, underscoring the substantial advantages conferred by unit constraints.
% Noise
While our method ranks second in noiseless scenarios, it excels in the presence of even minimal noise levels (exceeding 0.1\%), outperforming all other baseline methods in exact symbolic recovery. Notably, even at a higher noise level of 10\%, when scores for all other methods drop below 20\%, \PhySO\ consistently maintains a score above 50\%.
% Discussion
It is essential to note that many SR exploration strategies are driven by accuracy with minimal constraints on symbolic arrangement. Nevertheless, the paths leading to optimal fit quality and those leading to ideal symbolic arrangement (perfect fit quality and exact symbolic recovery) are not necessarily aligned. Essentially, one can enhance the fit quality of candidates over learning iterations while deviating further from the correct solution in terms of symbolic arrangement. As such, we attribute \PhySO's performance to the invaluable constraints on symbolic arrangements given by dimensional analysis.

\paragraph{Fit quality}
% Accuracy vs complexity concern
When considering the fraction of expressions with $R^2>0.999$  (defined as $R^2 = 1-{\sum_{i=1}^N (y_i - f(\mathbf{x}_i))^2}/{\sum_{i=1}^N (y_i - \bar{y})^2}$), numerous methods attain high scores by incorporating a substantial number of free constants. This approach often leads to the creation of complex expressions that frequently lack interpretability and deviate from dimensional analysis principles. While this may not pose significant issues in many fields in which accuracy is the only priority, it becomes a crucial concern in the context of the physical sciences. 
% Accuracy normalized by complexity
We therefore present in Figure \ref{fig:feynman_benchmark} the rate of accurate expressions normalized by the number of free constants. This metric provides insight into the efficiency of models while taking into account their complexity. In this context, \PhySO\ stands out as the leading method, excelling in the generation of concise, physically consistent, and interpretable expressions that approximate datasets, that is when it does not actually find the exact target symbolic expression.

\paragraph{Limitations}
It should be noted that our system requires all variables and constants involved to have defined physical units to exploit dimensional analysis constraints and that it is mostly adapted for exact symbolic recovery of concise and interpretable expressions. This largely limits its relevance to the physical sciences, in contrast to other SR methods that aim to produce longer but more accurate approximations that are computationally less expensive to evaluate than neural networks.
In addition, we note that contrary to end-to-end supervised learning based methods such as \cite{Kamienny_EndToEndSR, NeSymReS_EndToEndSR}, in the trial-and-error based strategy adopted here the neural network is reinitialized at the beginning of each SR task, preventing it from learning more generalized relations between datasets and analytical expressions. Future work will focus on equipping our framework with this ability as in \cite{uDSR_DSRbased}.

%\begin{figure}
%\includegraphics[angle=0, clip, width=0.5\hsize]{feynman_benchmark_pareto.pdf}
%\caption{Complexity versus rate of expressions having $R^2>0.999$ at a $10\%$ noise level for \texttt{PhySO} and other symbolic regression methods from the literature on the Feynman benchmark. \texttt{PhySO} is a Pareto optimum producing simple yet effective expressions. }
%\label{fig:feynman_benchmark_pareto}
%\end{figure}

\section{Conclusion}
\label{sec:conclusion}

We presented a Physical Symbolic Optimization (\PhySO) framework that is able to constrain the generation of equations on the fly to ensure that the rules of dimensional analysis are obeyed by construction. We expect this framework to be useful for many symbolic computation applications in the physical sciences. Here, we combined it with the state-of-the-art reinforcement learning strategies from \cite{PetersenDSR, SR_PG_improvements} resulting in a physics relevant SR method achieving state-of-the-art results on the standard Feynman benchmark from \texttt{SRBench}.

%\section*{Code availability}
%\label{sec:code}

\section*{Code availability}
\label{sec:code}

The documented code for the \PhySO\ algorithm, along with demonstration notebooks, is accessible on GitHub at \href{https://github.com/WassimTenachi/PhySO}{github.com/WassimTenachi/PhySO} \github{https://github.com/WassimTenachi/PhySO}, complete with comprehensive documentation. A frozen version related to this work is released under tag \href{https://github.com/WassimTenachi/PhySO/releases/tag/v1.0.0}{v1.0.0} \github{https://github.com/WassimTenachi/PhySO/releases/tag/v1.0.0} and deposited on zenodo: \href{https://doi.org/10.5281/zenodo.8415435}{10.5281/zenodo.8415435}.

For the sake of result reproducibility, we offer a straightforward method to replicate the outcomes presented in Figure \ref{fig:feynman_benchmark} by simply executing the following command: \texttt{python feynman\_run.py --equation i --noise n}. This command will run \texttt{PhySO} on challenge number \texttt{i} $\in \{0, 1, ..., 119\}$ of the Feynman benchmark, employing a noise level of \texttt{n} $\in [0,1]$.

Further enhancements and additional features for our software will continuously be updated and made accessible at \href{https://github.com/WassimTenachi/PhySO}{github.com/WassimTenachi/PhySO} \github{https://github.com/WassimTenachi/PhySO}, where interested users can stay abreast of the latest developments in our \PhySO\ framework.

\begin{ack}
RI acknowledges funding from the European Research Council (ERC) under the European Unions Horizon 2020 research and innovation programme (grant agreement No. 834148).
The authors would like to acknowledge the High Performance Computing Center of the University of Strasbourg for supporting this work by providing scientific support and access to computing resources. Part of the computing resources were funded by the Equipex Equip@Meso project (Programme Investissements d'Avenir) and the CPER Alsacalcul/Big Data.
\end{ack}

\bibliographystyle{plain}
\bibliography{refs}

%%%%%%%%%%%%%%%%%%%%%%%%%%%%%%%%%%%%%%%%%%%%%%%%%%%%%%%%%%%%

\end{document}